\documentclass[letterpaper, 10 pt, conference]{ieeeconf}  

\IEEEoverridecommandlockouts                              

\usepackage{graphicx} 
\usepackage{epsfig} 
\usepackage{mathptmx} 
\usepackage{times} 
\usepackage{amsmath} 
\usepackage{amssymb}  
\usepackage{caption}
\usepackage{subcaption}
\usepackage{graphicx}
\usepackage{url}
\usepackage[symbol]{footmisc}
\usepackage{hyperref}
\usepackage{cleveref}

\title{\LARGE \bf
PokeFlex: Towards a Real-World Dataset of Deformable Objects for Robotic Manipulation}

\author{Jan Obrist$^{1}$, Miguel Zamora$^{1}$, Hehui Zheng $^{2,3}$, Juan Zarate$^{4}$, Robert K. Katzschmann$^{2}$ and Stelian Coros$^{1}$
\thanks{$^{1}$Computational Robotics Lab, ETH Zurich, \{jobrist, mimora, scoros\}@ethz.ch}%
\thanks{$^{2}$Soft Robotics Lab, ETH Zurich, \{zhengh, rkk\}@ethz.ch}%
\thanks{$^{3}$ETH AI Center, ETH Zurich}%
\thanks{$^{4}$Advanced Interactive Technologies Lab, ETH Zurich, juan.zarate@ethz.ch}%
}

\begin{document}

\makeatletter
\let\@oldmaketitle\@maketitle
\renewcommand{\@maketitle}{
  \@oldmaketitle
  \begin{center}
  \vspace{10pt}
    \includegraphics[width=0.95\linewidth]{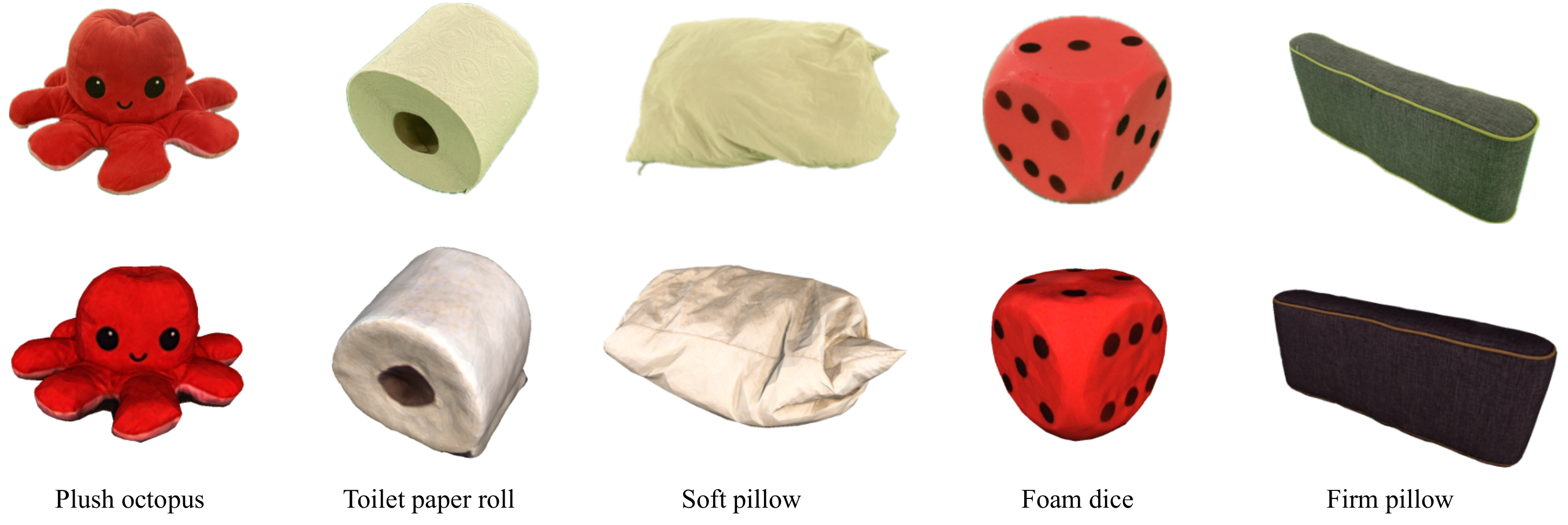}
    \captionsetup{font=footnotesize}
    \captionof{figure}{The PokeFlex dataset: Five objects shown as cropped RGB images (top) and 3D reconstructed meshes with texture (bottom).}
    \label{fig:objects}
    \vspace{-10pt}
  \end{center}\bigskip
}
\makeatother
\maketitle
\addtocounter{figure}{-1}
\thispagestyle{empty}
\pagestyle{empty}

\begin{abstract}
Advancing robotic manipulation of deformable objects can enable automation of repetitive tasks across multiple industries, from food processing to textiles and healthcare. Yet robots struggle with the high dimensionality of deformable objects and their complex dynamics. While data-driven methods have shown potential for solving manipulation tasks, their application in the domain of deformable objects has been constrained by the lack of data. To address this, we propose PokeFlex, a pilot dataset featuring real-world 3D mesh data of actively deformed objects, together with the corresponding forces and torques applied by a robotic arm, using a simple poking strategy. Deformations are captured with a professional volumetric capture system that allows for complete 360-degree reconstruction. The PokeFlex dataset consists of five deformable objects with varying stiffness and shapes. Additionally, we leverage the PokeFlex dataset to train a vision model for online 3D mesh reconstruction from a single image and a template mesh. We refer readers to the supplementary material and our \href{https://pokeflex-dataset.github.io/}{website}\footnote[1]{\url{https://pokeflex-dataset.github.io/}}  for demos and examples of our dataset.

\end{abstract}

\section{INTRODUCTION}

\begin{figure}[b!]
\centering
\includegraphics[width=\linewidth]{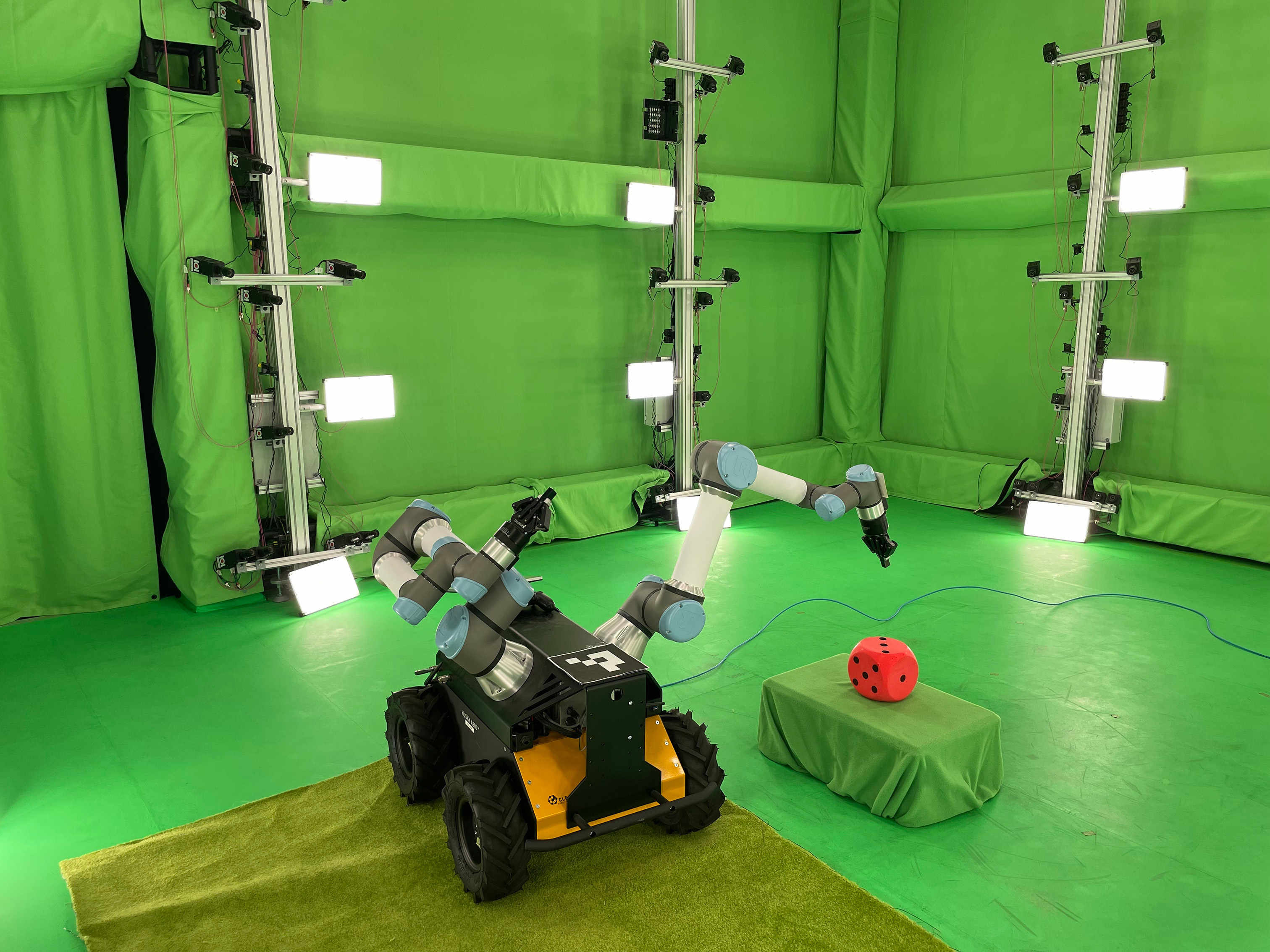}
\vspace{-0.45cm}
\captionsetup{font=footnotesize}
\caption{Setup for recording PokeFlex dataset: Husky dual arm robot is placed in a volumetric capture system within reach of the object. While the husky dual arm robot pokes the object, the deformations are recorded from the capture system at 30 fps. The robot records the position of the end effector and its acting forces and torques at 100 Hz.}
\label{fig:setup}
\end{figure}

Robotic manipulation of deformable objects is a relevant open challenge, as they are present in various settings ranging from industries to household environments. The challenge arises partly due to the objects' variable shapes, sizes, and complex material properties such as elasticity and plasticity. Recently, data-driven methods have achieved impressive results advancing the field of deformable object manipulation \cite{avigal2022speedfolding,yan2021learning}. However, there is limited data on deformable objects in current research. 

Such data can be beneficial for learning manipulation policies, estimating material parameters, and training 3D mesh reconstruction models. The latter is particularly relevant to enable the real-world deployment of traditional control methods based on mesh simulations. 

Previous datasets on deformable objects are limited to synthetic data \cite{huang2022defgraspsim}, RGB-D images and 3D mesh models of household objects in static deformed configurations without active manipulation \cite{garcia2022household}, or multi-view recordings and 3D point clouds of plush toys deformed with air-streams \cite{chen2022virtual}.

Our work proposes the PokeFlex dataset to address these gaps, capturing the real-world behavior of 5 deformable objects undergoing actively applied local deformations. The dataset includes 3D reconstructions of deformed meshes and the corresponding forces and torques induced by a robotic manipulator executing a simple poking strategy. The 3D meshes are reconstructed using a professional volumetric capture system \cite{volumetricCaptureSystem} with 106 cameras (RGB and infrared), allowing detailed 360\textdegree\,captures of the objects (\Cref{fig:setup}). 



\begin{figure}[t!]
  \begin{subfigure}{0.33\linewidth}
    \includegraphics[width=\linewidth]{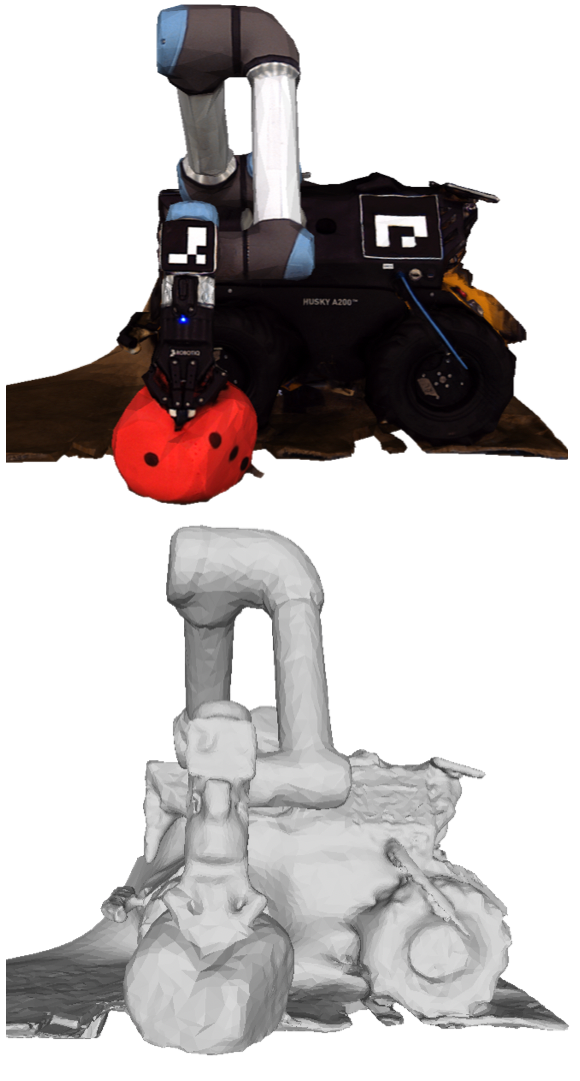}
    \caption{}
  \end{subfigure}%
  \hspace*{\fill}   
  \begin{subfigure}{0.33\linewidth}
    \includegraphics[width=\linewidth]{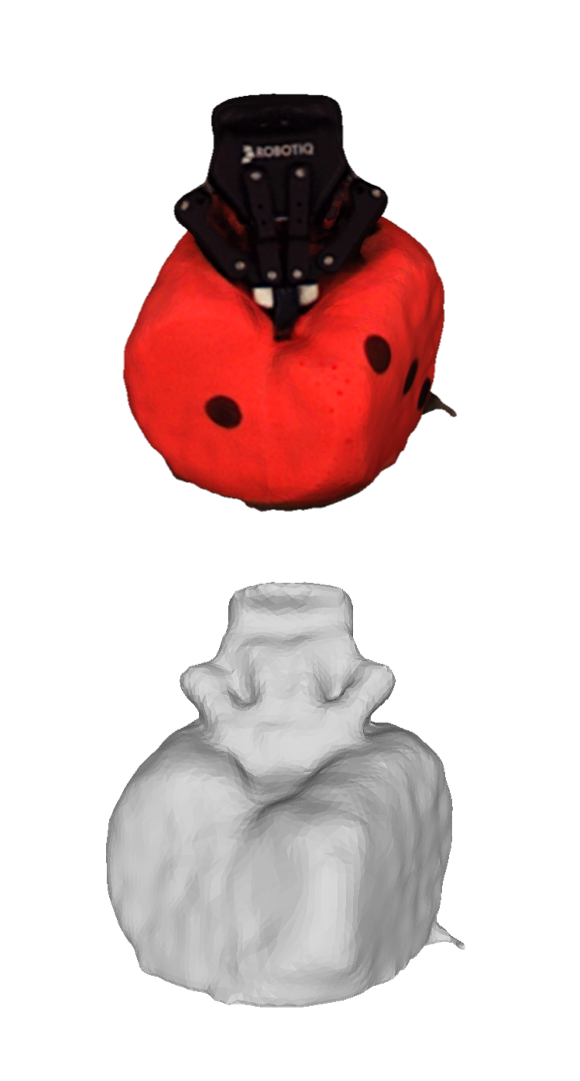}
    \caption{}
  \end{subfigure}%
  \hspace*{\fill}   
  \begin{subfigure}{0.33\linewidth}
    \includegraphics[width=\linewidth]{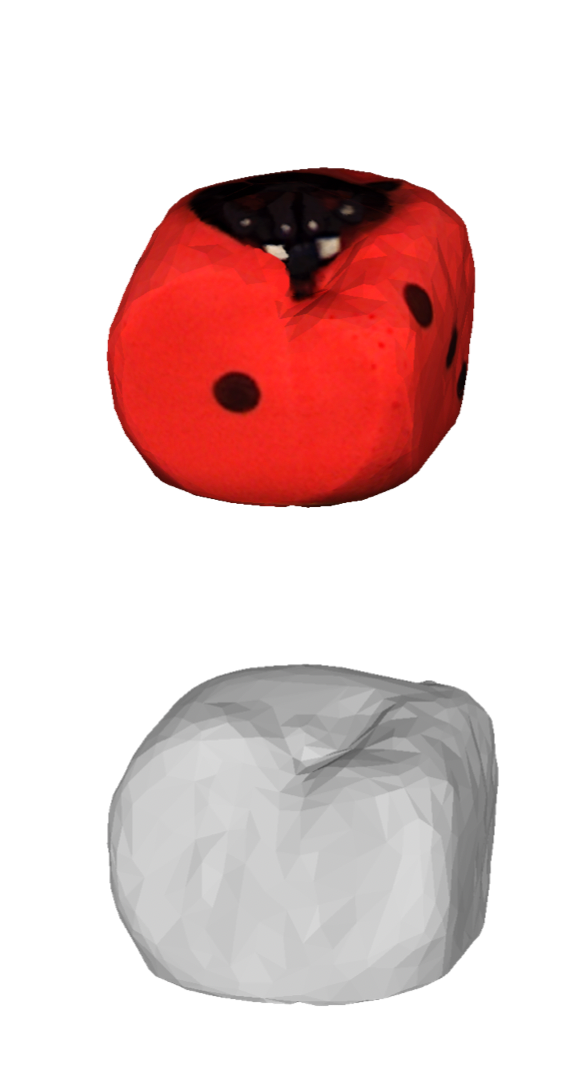}
    \caption{}
  \end{subfigure}
\captionsetup{font=footnotesize}
\caption{Two step mesh postprocessing: (a)$\rightarrow$(b) Using vertical and horizontal plane for clipping to remove most parts of the robotic system. (b)$\rightarrow$(c) Using frame based mesh clipping with 3D end effector model as mask.} \label{fig:postprocess}
\end{figure}

\begin{figure}[h!]
\centering
\vspace{0.3cm}
\includegraphics[width=0.93\linewidth]{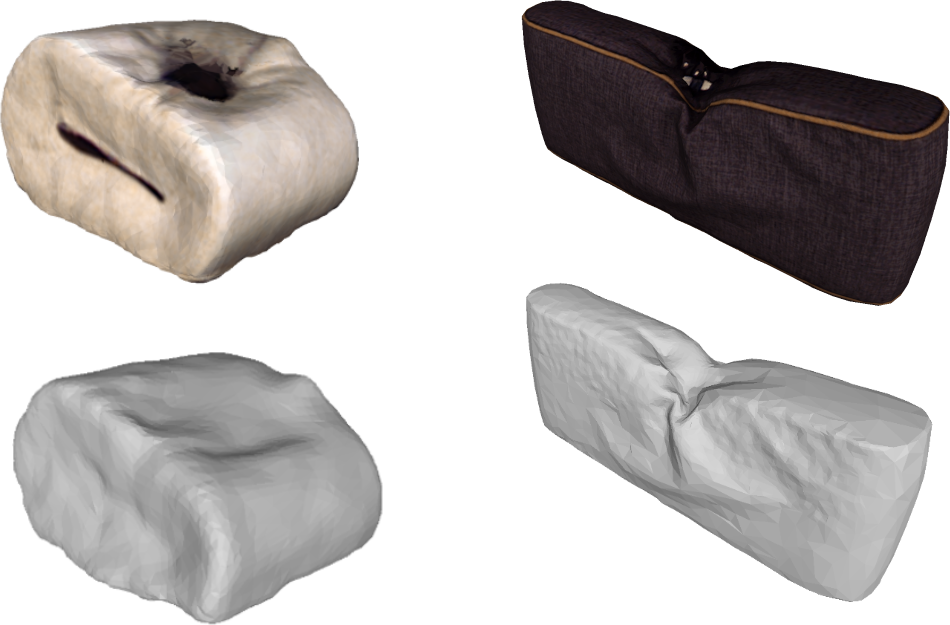}
\vspace{0.3cm}
\captionsetup{font=footnotesize}
\caption{Examples of reconstructed 3D meshes with and without texture (top and bottom respectively) for toilet paper roll (left) and firm pillow (right).}
\label{fig:deformations}
\end{figure}

\section{RESULTS}
\begin{figure}[h!]
\centering
\vspace{0.3cm}
\includegraphics[width=0.93\linewidth]{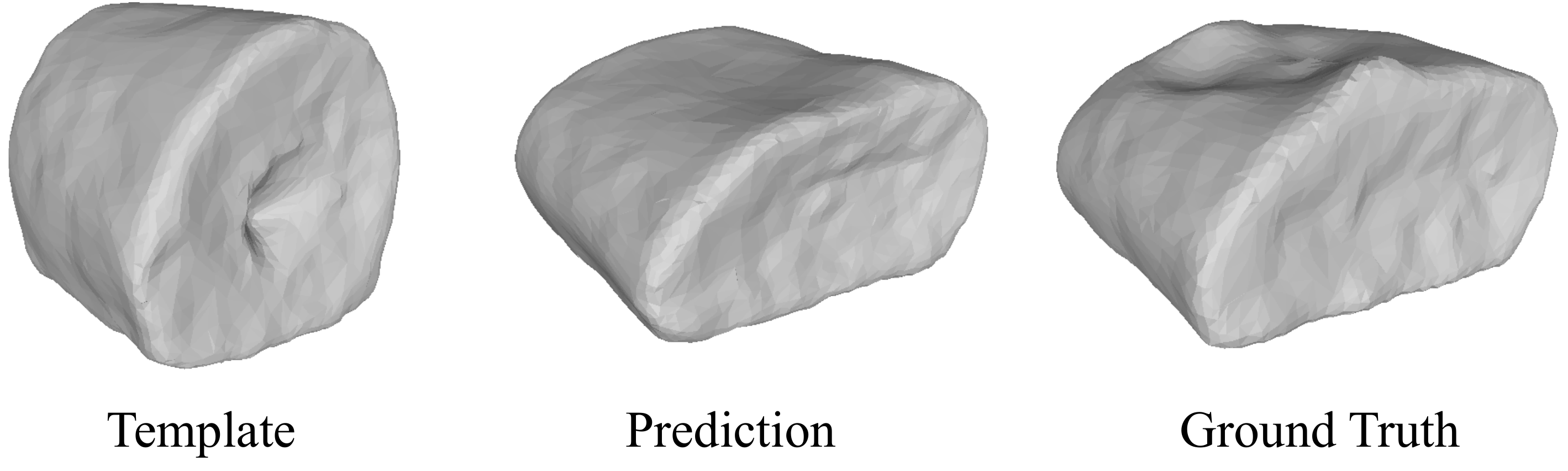}
\captionsetup{font=footnotesize}
\caption{Example prediction for the toilet paper roll object: Template mesh (left), predicted deformation (middle) and ground truth deformation (right).}
\label{fig:predictions}
\end{figure}

Our reconstruction pipeline, using a professional volumetric capture system \cite{volumetricCaptureSystem},  can effectively generate highly detailed 3D surface meshes for the five objects featured in the pilot of the PokeFlex dataset (\Cref{fig:objects}). 
For each of the featured objects, the current dataset includes 800 to 1,000 frames capturing their deformed configurations as exemplified in \Cref{fig:deformations}. Furthermore, each frame contains the following information:
\begin{itemize}
    \item 3D mesh model of the deformation
    \item 3D template mesh model
    \item Acting 3D forces and 3D torques.
    \item End-effector pose
    \item Camera recordings from the capture system
\end{itemize}
To validate the quality of the dataset, we extend the work of Mansour et al. \cite{mansour2024fast} to develop a method capable of predicting mesh deformations online. Previous methods that rely on point clouds to predict deformations are mainly trained on synthetic data \cite{mansour2024fast, Niemeyer_2019_ICCV, Lei_2022_CVPR}. However, real point cloud measurements are often noisy and sparse, leading to a sim-to-real gap. Other approaches using single images as input are not designed for online inference \cite{jack2018learning, kanazawa2018learning, wang2021pixel}. In our work, we use a Real-NVP model that allows predicting new vertex positions online from a single image and a template mesh. Preliminary experiments show promising results with an inference rate of 125 Hz (AMD Ryzen 7900 x 12 Core Processor CPU, NVIDIA GeForce RTX 4090 GPU with 24GB memory). \Cref{fig:predictions} shows that our model can predict the general deformation for the toilet paper roll object using a single image as input.

\section{CONCLUSIONS AND FUTURE WORK}
The preliminary 3D deformation prediction results (\Cref{fig:predictions}) showcase the quality of this pilot dataset. Further improvements in the accuracy of the deformation prediction can potentially be obtained by leveraging other data modalities such as the 3D forces and 3D torques present in the dataset. 

To encourage the adoption of the PokeFlex dataset by the community, the dataset will be extended to include 3D-printed deformable objects thus enhancing the reproducibility of our results with the release of the corresponding print files. Furthermore, the diversity of deformations applied to the objects will be improved by leveraging additional manipulation strategies such as pinching, dual arm squeezing, lifting, shaking, and tossing. 

We consider the PokeFlex dataset has the potential to advance the research on deformable objects, enabling a wide range of applications going from online 3D mesh reconstruction, to material parameter identification and policy learning for manipulation tasks. We look forward to making this dataset available for the community.


\section*{ACKNOWLEDGMENT}
This article is supported by the SDSC Grant entitled 'C22-08: Data-Driven Inference of Mesh-based Representations for Deformable Objects from Unstructured Point Clouds'

\addtolength{\textheight}{-17cm}   




\bibliographystyle{IEEEtran}
\bibliography{root.bib}

\end{document}